\documentclass[11pt,a4paper]{article}
\usepackage[hyperref]{acl2021}
\usepackage{times}
\usepackage{latexsym}

\usepackage[utf8]{inputenc}

\usepackage{microtype}

\aclfinalcopy % Uncomment this line for the final submission
\usepackage[linesnumbered,ruled]{algorithm2e}
\usepackage{graphicx}
\usepackage{subcaption}
\usepackage{xcolor}
\usepackage{bm}
\usepackage{amsfonts}
\usepackage{amsmath}
\usepackage{url}
\usepackage{siunitx}
\usepackage{booktabs}
\usepackage{multirow}
\usepackage{enumitem}
\newcommand{\norm}[1]{\left\lVert#1\right\rVert}

\title{Adversarial Training for Machine Reading Comprehension with Virtual Embeddings}

\author{Ziqing Yang$^1$, Yiming Cui$^{2,1}$, Chenglei Si$^3$,Wanxiang Che$^2$, \\
\textbf { Ting Liu$^2$, Shijin Wang$^{1,4}$,Guoping Hu$^{1}$}\\ 
{$^1$State Key Laboratory of Cognitive Intelligence, iFLYTEK Research, China}\\
{$^2$ Research Center for SCIR,} {Harbin Institute of Technology, Harbin, China}\\
{$^3$ University of Maryland,} {College Park MD, USA} \\
{$^4$iFLYTEK AI Research (Hebei), Langfang, China}\\
{\tt$^{1,4}$\{zqyang5,ymcui,sjwang3,gphu\}@iflytek.com}\\
{\tt $^2$\{ymcui,car,tliu\}@ir.hit.edu.cn},\ \ {\tt $^3$clsi@terpmail.umd.edu}\\
}

\date{}

\begin{document}
\maketitle
\begin{abstract}
Adversarial training (AT) as a regularization method has proved its effectiveness on various tasks. Though there are successful applications of AT on some NLP tasks, the distinguishing characteristics of NLP tasks have not been exploited. In this paper, we aim to apply AT on machine reading comprehension (MRC) tasks. Furthermore, we adapt AT for MRC tasks by proposing a novel adversarial training method called PQAT that perturbs the embedding matrix instead of word vectors. To differentiate the roles of passages and questions,  PQAT uses additional virtual P/Q-embedding matrices to gather the global perturbations of words from passages and questions separately. We test the method on a wide range of  MRC tasks, including span-based extractive RC and multiple-choice RC. The results show that adversarial training is effective universally, and PQAT further improves the performance.

\end{abstract}

\section{Introduction}

Neural networks have achieved superior performance on many tasks, but they are vulnerable to {\it adversarial examples} \cite{DBLP:journals/corr/SzegedyZSBEGF13} -- examples that have been mixed with certain perturbations.
{\em Adversarial training} (AT)  \cite{Goodfellow:2015} uses both clean and adversarial examples to improve the robustness of the model for image classification.

In the field of NLP, \citet{DBLP:conf/iclr/MiyatoDG17} have applied adversarial training on text classification tasks and improved the model performance. From then on,
many AT methods has been proposed \cite{wu-etal-2017-adversarial,yasunaga-etal-2018-robust,bekoulis-etal-2018-adversarial,DBLP:freelb,DBLP:smart,DBLP:alice,DBLP:alum}.
They mostly adopt a general AT strategy, but focus less on the adaptation of AT to NLP tasks. 
To explore this adaptation, in this work, we aim to apply adversarial training on machine reading comprehension (MRC) tasks, which exhibit complex NLP characteristics.

The objective of  MRC is to let a machine read the given passages and ask it to answer the related questions.
There are several types of MRC tasks. In this work we focus on span-based extractive RC \cite{rajpurkar-etal-2016-squad,rajpurkar-etal-2018-know,DBLP:conf/emnlp/Yang0ZBCSM18} and multiple-choice RC \cite{lai-etal-2017-race}.
To apply adversarial training on MRC tasks, we notice that there are several salient characteristics of MRC compared to other tasks such as image classification:
(1) The inputs are discrete. Unlike pixels, which can take continuous values, words are discrete tokens.
(2) The tokens in the input sequences are not independent. A word may occur in an input sequence several times. After the embedding layer, these occurrences are represented by the word vectors with the same value and hold the same semantic meaning (although the word may be polysemous).
(3) The roles of passages and questions are different. Given a question as the query, the model needs to look up the correct answer in the passage.

 \begin{table}[tbp]
 \small
\begin{tabular}{p{0.95\columnwidth}}
\toprule
{\bf Passage}: {\it ... The {\color{red} \textbf {rock}} cycle is an important concept in geology which illustrates the relationships between these three types of {\color{red} \textbf{rock}}, and magma. When a {\color{red} \textbf{rock}} crystallizes from melt (magma and/or lava), it is an {\color{orange} \textbf{igneous}}  {\color{red} \textbf{rock}}. ... } \\ \midrule
{\bf Question}: {\it An {\color{gray} \textbf{igneous}} {\color{blue} \textbf{rock}} is a {\color{blue} \textbf{rock}} that crystallizes from what?} \\
\bottomrule 
\end{tabular}

\caption{
An example from the SQuAD dataset. We highlight two words \emph{rock} and \emph{igneous} for better demonstration.
The words with the same color are injected with the same perturbation by PQAT.
The different occurrences of the same word (for example, \emph{rock} in passage and question) are perturbed differently depending on their roles.
}
\label{table:example}
\end{table}

People have utilized the first characteristic to apply adversarial training by perturbing input word vectors instead of tokens. However, the second and third characteristics have been largely ignored.
For example, in Table \ref{table:example}, which is a passage-question pair from the SQuAD dataset, the word \emph{rock}  has appeared multiple times. In the standard adversarial training, the perturbations added to each occurrence of \emph{rock} are different, ignoring the fact that they share the same meaning. 
On the other hand, the multiple occurrences of the same word in the passage and question play different roles, such as the \emph{rock} in the passage and question. It is appropriate to treat them differently.

To take the second and the third characteristics into consideration, we propose a novel adversarial training method called \textbf{PQAT}.
The core of PQAT is the virtual P/Q-embeddings, which are two independent embedding spaces for passages and questions. 
Each time we calculate perturbations, P/Q-embeddings gather the perturbations from passages and questions for each word, then generate a global and role-aware perturbation for each word from passages and questions separately. 
For example, in Table \ref{table:example}, the perturbations on all the occurrences of \emph{rock} in the passage and question will be gathered into two matrices separately, forming global and role-aware perturbations of \emph{rock}. PQAT is as efficient as the standard AT with nearly no extra time cost. Also, The virtual P/Q-embeddings are only used during training. They are discarded once the training is finished. Thus PQAT does not increase the model size and inference time for predictions.

We have applied adversarial training on several MRC tasks, including span-based extractive RC and multiple-choice RC.
Results show that adversarial training improves the MRC model performance universally and consistently, even over the strong pre-trained model baseline.
Furthermore, the PQAT method outperforms the standard AT on both normal datasets and adversarial datasets.
Lastly, our results verify the usefulness of incorporating information of task form into the design of the adversarial training method.

\section{Standard Adversarial Training}
Adversarial training first constructs adversarial examples by generating worst-case perturbations that maximize the current loss,  
then minimize the loss on those adversarial examples. 
In NLP tasks, a popular approach to generate perturbations is to perturb word vectors from the embedding layer \cite{DBLP:conf/iclr/MiyatoDG17}.
We denote the input token sequence as $X$ and the operation of looking up in an embedding layer $\bm{E}$ as  $emb(\bm{E},\cdot)$.
The objective of AT is 
 \begin{equation}
\underset{\theta, E}{\min}\ \mathbb{E}_{(X,y)\sim \mathcal{D}}\left[\underset{\norm{\delta}<\epsilon} {\max }\ \mathcal{L}(f_{\theta}(X_{vec}+\delta), y) \right]
\end{equation}
where $f_{\theta}(\cdot)$ is the model  parametrized by $\theta$ excluding word embedding layer;
 $X_{vec} = emb(\bm{E}, X) $ is the word vectors of input sequence.
$\mathcal{L}$ is the loss function. We perturb the word vectors with the adversarial perturbations $\delta$.

$\delta$ can be estimated by linearizing $\mathcal{L}(f_{\theta}(X_{vec}+\delta),y)$ around $X$
and perform the multiple-step projected gradient descent (PGD) \cite{DBLP:conf/iclr/MadryMSTV18}: 
\begin{align}
& \delta_{t+1} = \Pi_{\norm{\delta}\le \epsilon}(\delta_t + \alpha g_t /\norm{g_t})\\
& g_t = \nabla_{\delta}\mathcal{L}(f_{\theta}(X_{vec}+\delta),y)|_{\delta=\delta_t}
\end{align}

\noindent where $t$ is the gradient descent step, 
$ \Pi_{\norm{\delta}\le \epsilon}$ denotes projection $\delta$ back onto the $\epsilon$-ball.
$g_t$  is the gradient of the loss with respect to perturbation $\delta$. 
The more gradient descent steps, the better approximation of $\delta$,  but also more expensive in computation.

\section{Adversarial Training for MRC}

\begin{figure}[tbp]
\centering
\includegraphics[width=\columnwidth]{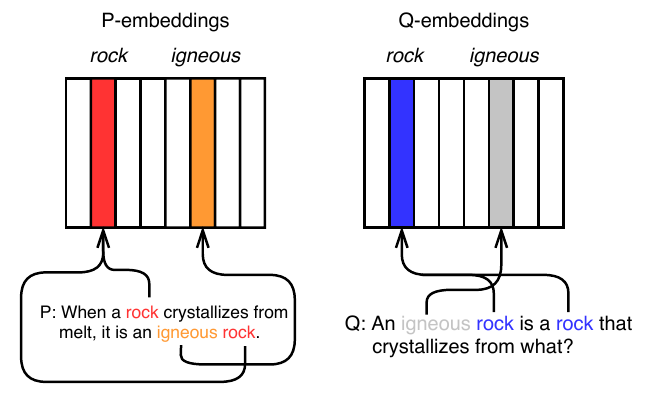}
\caption{P/Q-emebddings collect the perturbations on each word from passages and questions separately.}
\label{fig:pq-embeddings}
\end{figure}

In the above algorithm,  
when generating the perturbations on $X_{vec}$ through backward propagation, each word vector $X_{vec}^i$ is perturbed independently, like the pixels in an image. 
It ignores the semantic relationship among the word vectors of a word's different occurrences. To make the perturbation on each occurrence aware of other occurrences of the same word, we adapt AT by gathering not only the perturbations on each word vector, but also the perturbations on the embedding matrix. 
The latter can be seen as the global perturbations, which provide context-insensitive semantic information.

The global perturbations are rather coarse-grained, since all the occurrences of the same word receive the same global perturbation.
Note that in MRC tasks, words in passages and questions play different roles. 
Thus, to keep this information, we distinguish the words in passages and questions by creating two \emph{virtual} embedding matrices $\bm{P}$ and $\bm{Q}$: P-embedding matrix $\bm{P}$ collects the perturbations of all the words from the passages; Q-embedding matrix $\bm{Q}$ for the questions. We give an illustration in Figure \ref{fig:pq-embeddings}. P/Q-embedding matrices are \emph{virtual} since they only provide perturbations, no the real word vectors. During training, perturbations from virtual embeddings and word vectors are summed up to form the adversarial input $Z_{vec}$. The final objective is
\begin{align}
& \underset{\theta, E}{\min}\ \mathbb{E}_{(X,y)\sim \mathcal{D}}\left[\underset{\norm{\delta}<\epsilon} {\max }\ \mathcal{L}(f_{\theta}(Z_{vec}), y) \right] \\
& Z_{vec} = [X_{vec}^{P} + P_{vec}; X_{vec}^{Q}+ Q_{vec}] + \delta
\end{align}
$P_{vec} = emb (\bm{P},X^{P})$, $Q_{vec} = emb (\bm{Q},X^{Q})$ are the perturbations from the virtual embeddings.
$X^{P}$ and $X^{Q}$ stand for the passage and question sections in $X$. $[\cdot ;\cdot]$ denotes concatenation.
In this way,  we have generated fine-grained local perturbations $\delta$ by standard AT, and global role-aware perturbations $P_{vec}$ and $Q_{vec}$ by the virtual P/Q-embeddings. We call the later process as \textbf{PQAT}, which is the main adaptation of adversarial training for MRC.

We list the overall algorithm of adversarial training for MRC in Algorithm \ref{algo:mrc-at}. We initialize $\bm{P}$ and $\bm{Q}$ with the gaussian distribution. For each batch, we perform $K$-step gradient descent (line 9--22): we look up the original word vectors and P/Q-embedding vectors from the embedding layer $\bm{E}$  and the P/Q-embedding matrices. The adversarial inputs are constructed by summing them with local perturbations $\delta$. Then we compute the gradients of model parameters $\bm{g_t}$, local perturbations $\bm{g_{\delta}}$ and P/Q-embedding matrices $\bm{g_P}$ and $\bm{g_Q}$. These gradients can be calculated in a single backward pass.
Lastly, we update the virtual embeddings and local perturbations (line 18--21). 

Note that P/Q-embedding matrices serve as the containers for perturbations. 
When the training is finished, P/Q-embedding matrices are no longer needed and can be discarded.

$\epsilon_{\delta}$,  $\epsilon_{P}$ and $\epsilon_{Q}$ control the strengths of standard AT and PQAT.
If $\epsilon_{\delta}=0$, we have a pure P/Q-embeddings based adversarial training, i.e., PQAT;  while if $\epsilon_{P}=\epsilon_{Q}=0$, we recover the standard AT.

%%%%%%%%%%%%%%--Algorithm start--%%%%%%%%%%%%%%%%%
\SetKwInput{KwNotation}{Notation}
\begin{algorithm}[t]
\small
\caption{Adversarial Training for \mbox{Machine} Reading Comprehension} 
\label{algo:mrc-at}
\KwNotation{
$V$ is the vocabulary size; $D$ is the embedding dimension.}
\KwIn{Training samples $ \mathcal{D}=\{(X,y)\}$, P/Q-embedding matrices $\bm{P},\bm{Q}\in \mathbb{R}^{V\times D}$,
initialization variance $\sigma$, perturbation strength $\{\epsilon_{\delta}, \epsilon_{P}, \epsilon_{Q}\}$, 
adversarial steps $K$.}
\emph{Initialize P/Q-embedding matrices}\\
  $\bm{P} \leftarrow \mathcal{N}(0, \sigma^2 I)$ ,   $\bm{Q} \leftarrow \mathcal{N}(0, \sigma^2 I)$ \\

  \For {batch $B \subset  \mathcal{D}$}
  { 
  \emph{\small Normalize P/Q-embedding matrices}\\
   $\bm{P} \leftarrow (\bm{P}-mean(\bm{P})/{std(\bm{P}) }\cdot \sigma$ \\
   $\bm{Q} \leftarrow (\bm{Q}-mean(\bm{Q}))/{std(\bm{Q})} \cdot \sigma$ \\
   \emph{\small Initialize perturbation and gradient} \\
   $\delta \leftarrow \frac{1}{\sqrt{D}} U(-\sigma, \sigma) $, $\bm{g_0} \leftarrow 0 $\\
   \For {$t = 1, \ldots, K$}
   {
    $X_{vec} = emb(\bm{E},X)$ \\
    $P_{vec} = emb(\bm{P},X^{P})$ \\
    $Q_{vec} = emb(\bm{Q},X^{Q})$ \\
     $Z_{vec} = X_{vec} + P_{vec} + Q_{vec} + \delta$ \\
   $\bm{g}_t = \bm{g}_{t-1} +\mathbb{E}[\nabla_{\theta,\bm{E}} \mathcal{L}(f_{\theta}(Z_{vec}), y)]$ \\
   $\bm{g_{\delta}} =  \mathbb{E}[\nabla_{\delta} \mathcal{L}(f_{\theta}(Z_{vec}), y)]$ \\
   $\bm{g_{P}} =  \mathbb{E}[\nabla_{\bm{P}} \mathcal{L}(f_{\theta}(Z_{vec}), y)]$ \\
   $\bm{g_{Q}} =  \mathbb{E}[\nabla_{\bm{Q}} \mathcal{L}(f_{\theta}(Z_{vec}), y)]$ \\
   
   $\delta \leftarrow \delta +  \bm{g_{\delta}}/  \norm{\bm{g_{\delta}}}_2 \cdot \norm{{X_{vec}}}_2 \epsilon_{\delta}$ \\
   \emph{Update with token-wise normalization} \\
   $P^i \leftarrow  P^i +\bm{g_{P}}^i / \norm{\bm{g_{P}}^i}_2 \cdot \norm{X_{vec}^i}_2 \epsilon_{P}$  \\
   $Q^i \leftarrow  Q^i + \bm{g_{Q}}^i / \norm{\bm{g_{Q}}^i}_2 \cdot \norm{X_{vec}^i }_2 \epsilon_{Q}$ 
   }
   $ \{\theta,\bm{E}\} \leftarrow \textrm{AdamUpdate}(\{\theta,\bm{E}\}, \bm{g}_K)$
  }

\end{algorithm}
%%%%%%%%%%%%%%--Algorithm end--%%%%%%%%%%%%%%%%%

%%%%%%%%%%%%%%TABLE%%%%%%%%%%%%%%
\begin{table*}[t]
   \centering
   \small
    \centering
    \begin{tabular}{@{}lccccccc@{}}
      \toprule
      \multirow{2}{*}{\textbf{Model}} & \multicolumn{2}{c}{\textbf{SQuAD 1.1}}                  & \multicolumn{2}{c}{\textbf{SQuAD 2.0}} & \multicolumn{2}{c}{\textbf{HotpotQA}}                  &\textbf{RACE} \\
                                 & \textbf{EM}              & \textbf{F1}              & \textbf{EM}              & \textbf{F1}         & \textbf{joint EM}              & \textbf{joint F1}              & \textbf{Acc}         \\  \midrule
      {\small \emph{BASE setting}}  & & & & \\
      RoBERTa & 84.72 & 91.54  & 79.77 & 83.18  & 41.70                 & 	 69.30  & 74.75\\ 

      PQAT & \textbf{85.87}                 & \textbf{92.33}                     & \textbf{81.66}            & \textbf{84.79}     &  \textbf{43.03} & 	\textbf{70.40}  & \textbf{76.32}       \\ \midrule \midrule
      {\small \emph{LARGE setting}}  & & & & \\
      RoBERTa  &87.76  & 	93.90 & 85.67 & 	88.86 & 45.91$^\dag$                 & 	 73.93$^\dag$  & 84.66 \\
      PQAT & \textbf{88.32}	 & \textbf{94.34} & \textbf{86.35} & 	\textbf{89.49} &  \textbf{46.79} & 	\textbf{74.63}  & \textbf{86.02} \\  \bottomrule
    \end{tabular}

    \label{table:main_sq}
  \caption{Results on the development sets of SQuAD 1.1, SQuAD 2.0 and HotpotQA, and results on the test set of RACE. 
  $^\dag$: the results are taken from \newcite{shao2020graph}.}
 \label{table:overall}
\end{table*}
%%%%%%%%%%%%%%TABLE%%%%%%%%%%%%%%

\section{Experiments Setup}
\textbf{Datasets}. We perform experiments on several English MRC tasks, including span-based extractive MRC tasks -- SQuAD 1.1 \cite{rajpurkar-etal-2016-squad}, SQuAD 2.0 \cite{rajpurkar-etal-2018-know}, HotpotQA \cite{DBLP:conf/emnlp/Yang0ZBCSM18}, and multiple-choice MRC task RACE \cite{lai-etal-2017-race}. We also test model robustness on the adversarial datasets {AddSent} and{AddOneSent} \cite{jia-liang-2017-adversarial}.

\noindent \textbf{Model Settings}. We build the  MRC model with RoBERTa \cite{DBLP:journals/corr/abs-1907-11692}, following the standard model structure for SQuAD and RACE \cite{devlin2018bert}. For HotpotQA,  we follow the model in \newcite{shao2020graph}. It uses RoBERTa as the encoder followed by a multi-task prediction layer.
We denote the passage as $P$ and the question as $Q$. 
To construct the inputs,  for span-based extractive RC, we concatenate each $P$ and $Q$ with model-dependent special tokens;
for multiple-choice RC with $m$ options for each example,  we append each option to the concatenation of $P$ and $Q$,  and construct $m$ input sequences from each example.

When applying AT or PQAT, we only perturb the word embeddings and leave the position embeddings unchanged. For PQAT on RACE, we let the Q-embedding matrix collect perturbations from both questions and options.

\noindent \textbf{Training Settings and Hyperparameters}. All the models are implemented with Transformers \cite{Wolf2019HuggingFacesTS} and trained on a single Nvidia V100 GPU.
To improve the stability and reduce the uncertainty of the results, we run each experiment four times with different seeds and report the mean value of performance. 
We use AdamW as our optimizer with batch size 24 and learning rate 3e-5 for RoBERTa$_{\texttt{BASE}}$  and  2e-5 or 1e-5 for RoBERTa$_{\texttt{LARGE}}$. 
The maximum number of epochs is set to $3$ for SQuAD and $5$ for RACE and HotpotQA. A linear learning rate decay schedule with warmup ratio 0.1 was used.
For PQAT,  $\epsilon_{\delta}$ is set to 0, $\epsilon_{P}$ and $\epsilon_{Q}$ is set to 4e-2 for RACE and 2e-2 for other tasks.  The variance  $\sigma$ is 
1e-2.
We set the number of gradient descent steps $K=2$ to balance speed and performance.

\section{Results}

\begin{table}[t]
  \centering
  \small
    \begin{tabular}{@{}lccc@{}}
       \toprule
      \multirow{2}{*}{\textbf{Model}} & {\textbf{SQuAD 1.1}}                  & {\textbf{SQuAD 2.0}} &{\textbf{RACE}}\\
                                 & \textbf{EM}                     & \textbf{EM}             & \textbf{Acc}             \\ \midrule
                                 \emph{BASE setting} &  & & \\
       \multirow{2}{*}{\ \ \ \ \underline{PQAT}}       & \underline{ {85.87}}{\ \ \ }                                 & \underline{\textbf{81.66}}{\ \ \ }       & \underline{{76.32}}{\ \ \ }            \\         
                                                                                         & (\it{0.08}){\ \ \ }                                       & (\it{0.21}){\ \ \ }           & (\it{0.32}){\ \ \ }         \\
       
       \multirow{2}{*}{\ \  \ \ PQAT + AT}             &\textbf{85.96}{\ \ \ }                       & 81.11 \color{red}{$\downarrow$}                    &\textbf{76.50}{\ \ \ }  \\   
                                                                                    & (\it{0.10}){\ \ \ }                                    & (\it{0.14}){\ \ \ }                    & (\it{0.35}){\ \ \ } \\
       \multirow{2}{*}{\ \  \ \ AT}   & 85.64 \color{red}{$\downarrow$}                   & 81.23 \color{red}{$\downarrow$}            & 75.94 \color{red}{$\downarrow$}     \\
                                                                                                        &(\it{0.15}){\ \ \ }                                   & (\it{0.30}){\ \ \ }    & (\it{0.37}){\ \ \ }       \\

    \bottomrule
    \end{tabular}
    \caption{Comparison of PQAT, standard AT and the combination. AT is short for Standard AT. Arrows indicate the drops relative to the \underline{PQAT}. Numbers in the parentheses are the standard deviations.}
    \label{table:ablation}
\end{table}

\begin{table}[th]
\small
  \begin{tabular}{lccc}
\toprule
\textbf{Model}  & \textbf{AddSent}     & \textbf{AddOneSent}   & \textbf{Dev}       \\ \midrule
R.M-Reader$^{\dag}$                             & 58.5                       & 67.0                                                                         & 86.6          \\
KAR$^{\ddag}$                                 & 60.1                       & 72.3                                                                    & 83.5             \\
ALUM$_{\texttt{BERT-BASE}}$$^{\S}$                                 & 60.4                      & 69.8                                                                    & 90.8             \\
RoBERTa$_{\texttt{BASE}}$ & 59.7 & 68.8 & 91.5  \\
 \midrule
PQAT                             & \bf{64.7}                       & \bf{73.6}                                                                     & \bf{92.3}      \\
AT     & 63.2 & 72.6 & 92.1 \\ 
\bottomrule
\end{tabular}
\caption{Model performance (F1) on AddSent, AddOneSent and SQuAD 1.1 dev set. AT is short for Standard AT. $^{\dag}$:\newcite{kar}. $^{\ddag}$: \newcite{rmr}, $^{\S}$: \newcite{DBLP:alum}.}
\label{table:robust}
\end{table}

\subsection{Overall Results}
The overall results are summarized in Table \ref{table:overall}, where we compare PQAT with the baseline. PQAT is able to boost model performance across all MRC tasks and outperforms the RoBERTa baseline significantly. On HotpotQA, which is a complicated MRC task that features multi-hop questions and asks for multiple kinds of predictions, PQAT still outperforms the baseline by  1.3/1.1 on Joint EM/Joint F1. On RACE, PQAT improves the performance significantly by 1.5\% in accuracy. The universal improvements on various kinds of MRC tasks prove the wide applicability of PQAT.

\subsection{Comparison}

We compare different adversarial training methods and their combinations by tuning the strengths of perturbations $\{\epsilon_{\delta},\epsilon_{P}, \epsilon_{Q}\}$.
The results are in Table \ref{table:ablation}. The underlined scores are the ones reported in Table \ref{table:overall}.
Firstly, to test the effectiveness of standard AT, we disable  PQAT with  $\epsilon_{P}=\epsilon_{Q}=0$ and enable standard AT with $\epsilon_{\delta}=$2e-3 for RACE and 1e-2 for other tasks \footnote{We have searched from 1e-3 to 1e-1 and taken the best value.}. Other settings are unchanged, and we still follow Algorithm \ref{algo:mrc-at}.
PQAT consistently outperforms standard AT on the three tasks. 
Then we enable both PQAT and standard AT by setting all the strengths $\{\epsilon_{\delta},\epsilon_{P}, \epsilon_{Q}\}$ to non-zero values. The performance gets slightly better on SQuAD 1.1 and RACE, but gets worse on SQuAD 2.0. 

Compared with the standard AT, PQAT achieves higher performance by itself. Therefore PQAT could be a better alternative to applying adversarial training on MRC tasks.

\subsection{Robustness on Adversarial Datasets}
We assess the robustness of MRC models with AddSent and AddOneSent.
{AddSent} and {AddOneSent} are two adversarial datasets built on SQuAD 1.1. In both datasets, passages are appended with distracting sentences.
 MRC models that heavily rely on text matching may be easily fooled to predict wrong answers from the distracting sentences.

The results are shown in Table \ref{table:robust}. With the standard adversarial training (AT), the MRC model improves its robustness by about 5\%   over RoBERTa$_\texttt{BASE}$  in F1. 
PQAT further improves the performance over AT by about 1\% on both AddSent and AddOneSent.

\section{Conclusion}
We have applied adversarial training on a wide range of  MRC tasks, including span-based extractive RC and multiple-choice RC.
Especially, we have proposed a novel adversarial training method PQAT, which uses virtual P/Q-embedding matrices to generate global and role-aware perturbations that consider the characteristics of MRC tasks.
Our experiments demonstrate that adversarial training improves the MRC model performance universally and consistently, even over the strong pre-trained model baseline.
 The PQAT method further improves the model performance over the standard AT on both normal datasets and adversarial datasets.

\section*{Acknowledgments}

We would like to thank all anonymous reviewers for their valuable comments on our work. 
This work is funded by National Key R\&D Program of China (No.2018YFC0831601).
\bibliography{mybib}

\begin{thebibliography}{22}
\expandafter\ifx\csname natexlab\endcsname\relax\def\natexlab#1{#1}\fi

\bibitem[{Bekoulis et~al.(2018)Bekoulis, Deleu, Demeester, and
  Develder}]{bekoulis-etal-2018-adversarial}
Giannis Bekoulis, Johannes Deleu, Thomas Demeester, and Chris Develder. 2018.
\newblock \href {https://www.aclweb.org/anthology/D18-1307} {Adversarial
  training for multi-context joint entity and relation extraction}.
\newblock In \emph{Proceedings of the 2018 Conference on Empirical Methods in
  Natural Language Processing}, pages 2830--2836, Brussels, Belgium.
  Association for Computational Linguistics.

\bibitem[{Devlin et~al.(2018)Devlin, Chang, Lee, and
  Toutanova}]{devlin2018bert}
Jacob Devlin, Ming-Wei Chang, Kenton Lee, and Kristina Toutanova. 2018.
\newblock Bert: Pre-training of deep bidirectional transformers for language
  understanding.
\newblock \emph{arXiv preprint arXiv:1810.04805}.

\bibitem[{Goodfellow et~al.(2015)Goodfellow, Shlens, and
  Szegedy}]{Goodfellow:2015}
Ian Goodfellow, Jonathon Shlens, and Christian Szegedy. 2015.
\newblock \href {http://arxiv.org/abs/1412.6572} {Explaining and harnessing
  adversarial examples}.
\newblock In \emph{International Conference on Learning Representations}.

\bibitem[{Hu et~al.(2018)Hu, Peng, Huang, Qiu, Wei, and Zhou}]{rmr}
Minghao Hu, Yuxing Peng, Zhen Huang, Xipeng Qiu, Furu Wei, and Ming Zhou. 2018.
\newblock \href {https://doi.org/10.24963/ijcai.2018/570} {Reinforced mnemonic
  reader for machine reading comprehension}.
\newblock In \emph{Proceedings of the Twenty-Seventh International Joint
  Conference on Artificial Intelligence, {IJCAI} 2018, July 13-19, 2018,
  Stockholm, Sweden.}, pages 4099--4106.

\bibitem[{Jia and Liang(2017)}]{jia-liang-2017-adversarial}
Robin Jia and Percy Liang. 2017.
\newblock \href {https://doi.org/10.18653/v1/D17-1215} {Adversarial examples
  for evaluating reading comprehension systems}.
\newblock In \emph{Proceedings of the 2017 Conference on Empirical Methods in
  Natural Language Processing}, pages 2021--2031, Copenhagen, Denmark.
  Association for Computational Linguistics.

\bibitem[{Jiang et~al.(2019)Jiang, He, Chen, Liu, Gao, and Zhao}]{DBLP:smart}
Haoming Jiang, Pengcheng He, Weizhu Chen, Xiaodong Liu, Jianfeng Gao, and Tuo
  Zhao. 2019.
\newblock \href {http://arxiv.org/abs/1911.03437} {{SMART:} robust and
  efficient fine-tuning for pre-trained natural language models through
  principled regularized optimization}.
\newblock \emph{CoRR}, abs/1911.03437.

\bibitem[{Lai et~al.(2017)Lai, Xie, Liu, Yang, and Hovy}]{lai-etal-2017-race}
Guokun Lai, Qizhe Xie, Hanxiao Liu, Yiming Yang, and Eduard Hovy. 2017.
\newblock \href {https://doi.org/10.18653/v1/D17-1082} {{RACE}: Large-scale
  {R}e{A}ding comprehension dataset from examinations}.
\newblock In \emph{Proceedings of the 2017 Conference on Empirical Methods in
  Natural Language Processing}, pages 785--794, Copenhagen, Denmark.
  Association for Computational Linguistics.

\bibitem[{Liu et~al.(2020)Liu, Cheng, He, Chen, Wang, Poon, and
  Gao}]{DBLP:alum}
Xiaodong Liu, Hao Cheng, Pengcheng He, Weizhu Chen, Yu~Wang, Hoifung Poon, and
  Jianfeng Gao. 2020.
\newblock \href {http://arxiv.org/abs/2004.08994} {Adversarial training for
  large neural language models}.
\newblock \emph{CoRR}, abs/2004.08994.

\bibitem[{Liu et~al.(2019)Liu, Ott, Goyal, Du, Joshi, Chen, Levy, Lewis,
  Zettlemoyer, and Stoyanov}]{DBLP:journals/corr/abs-1907-11692}
Yinhan Liu, Myle Ott, Naman Goyal, Jingfei Du, Mandar Joshi, Danqi Chen, Omer
  Levy, Mike Lewis, Luke Zettlemoyer, and Veselin Stoyanov. 2019.
\newblock \href {http://arxiv.org/abs/1907.11692} {Roberta: {A} robustly
  optimized {BERT} pretraining approach}.
\newblock \emph{CoRR}, abs/1907.11692.

\bibitem[{Madry et~al.(2018)Madry, Makelov, Schmidt, Tsipras, and
  Vladu}]{DBLP:conf/iclr/MadryMSTV18}
Aleksander Madry, Aleksandar Makelov, Ludwig Schmidt, Dimitris Tsipras, and
  Adrian Vladu. 2018.
\newblock \href {https://openreview.net/forum?id=rJzIBfZAb} {Towards deep
  learning models resistant to adversarial attacks}.
\newblock In \emph{6th International Conference on Learning Representations,
  {ICLR} 2018, Vancouver, BC, Canada, April 30 - May 3, 2018, Conference Track
  Proceedings}. OpenReview.net.

\bibitem[{Miyato et~al.(2017)Miyato, Dai, and
  Goodfellow}]{DBLP:conf/iclr/MiyatoDG17}
Takeru Miyato, Andrew~M. Dai, and Ian~J. Goodfellow. 2017.
\newblock \href {https://openreview.net/forum?id=r1X3g2\_xl} {Adversarial
  training methods for semi-supervised text classification}.
\newblock In \emph{5th International Conference on Learning Representations,
  {ICLR} 2017, Toulon, France, April 24-26, 2017, Conference Track
  Proceedings}.

\bibitem[{Pereira et~al.(2020)Pereira, Liu, Cheng, Asahara, and
  Kobayashi}]{DBLP:alice}
Lis Pereira, Xiaodong Liu, Fei Cheng, Masayuki Asahara, and Ichiro Kobayashi.
  2020.
\newblock \href {http://arxiv.org/abs/2005.08156} {Adversarial training for
  commonsense inference}.
\newblock \emph{CoRR}, abs/2005.08156.

\bibitem[{Rajpurkar et~al.(2018)Rajpurkar, Jia, and
  Liang}]{rajpurkar-etal-2018-know}
Pranav Rajpurkar, Robin Jia, and Percy Liang. 2018.
\newblock \href {https://www.aclweb.org/anthology/P18-2124} {Know what you
  don{'}t know: Unanswerable questions for {SQ}u{AD}}.
\newblock In \emph{Proceedings of the 56th Annual Meeting of the Association
  for Computational Linguistics (Volume 2: Short Papers)}, pages 784--789,
  Melbourne, Australia. Association for Computational Linguistics.

\bibitem[{Rajpurkar et~al.(2016)Rajpurkar, Zhang, Lopyrev, and
  Liang}]{rajpurkar-etal-2016-squad}
Pranav Rajpurkar, Jian Zhang, Konstantin Lopyrev, and Percy Liang. 2016.
\newblock \href {https://doi.org/10.18653/v1/D16-1264} {{SQ}u{AD}: 100,000+
  questions for machine comprehension of text}.
\newblock In \emph{Proceedings of the 2016 Conference on Empirical Methods in
  Natural Language Processing}, pages 2383--2392, Austin, Texas. Association
  for Computational Linguistics.

\bibitem[{Shao et~al.(2020)Shao, Cui, Liu, Wang, and Hu}]{shao2020graph}
Nan Shao, Yiming Cui, Ting Liu, Shijin Wang, and Guoping Hu. 2020.
\newblock \href {https://www.aclweb.org/anthology/2020.emnlp-main.583} {Is
  {G}raph {S}tructure {N}ecessary for {M}ulti-hop {Q}uestion {A}nswering?}
\newblock In \emph{Proceedings of the 2020 Conference on Empirical Methods in
  Natural Language Processing (EMNLP)}, pages 7187--7192, Online. Association
  for Computational Linguistics.

\bibitem[{Szegedy et~al.(2014)Szegedy, Zaremba, Sutskever, Bruna, Erhan,
  Goodfellow, and Fergus}]{DBLP:journals/corr/SzegedyZSBEGF13}
Christian Szegedy, Wojciech Zaremba, Ilya Sutskever, Joan Bruna, Dumitru Erhan,
  Ian~J. Goodfellow, and Rob Fergus. 2014.
\newblock \href {http://arxiv.org/abs/1312.6199} {Intriguing properties of
  neural networks}.
\newblock In \emph{2nd International Conference on Learning Representations,
  {ICLR} 2014, Banff, AB, Canada, April 14-16, 2014, Conference Track
  Proceedings}.

\bibitem[{Wang and Jiang(2018)}]{kar}
Chao Wang and Hui Jiang. 2018.
\newblock \href {http://arxiv.org/abs/1809.03449} {Exploring machine reading
  comprehension with explicit knowledge}.
\newblock \emph{CoRR}, abs/1809.03449.

\bibitem[{Wolf et~al.(2019)Wolf, Debut, Sanh, Chaumond, Delangue, Moi, Cistac,
  Rault, Louf, Funtowicz, and Brew}]{Wolf2019HuggingFacesTS}
Thomas Wolf, Lysandre Debut, Victor Sanh, Julien Chaumond, Clement Delangue,
  Anthony Moi, Pierric Cistac, Tim Rault, R'emi Louf, Morgan Funtowicz, and
  Jamie Brew. 2019.
\newblock Huggingface's transformers: State-of-the-art natural language
  processing.
\newblock \emph{ArXiv}, abs/1910.03771.

\bibitem[{Wu et~al.(2017)Wu, Bamman, and Russell}]{wu-etal-2017-adversarial}
Yi~Wu, David Bamman, and Stuart Russell. 2017.
\newblock \href {https://doi.org/10.18653/v1/D17-1187} {Adversarial training
  for relation extraction}.
\newblock In \emph{Proceedings of the 2017 Conference on Empirical Methods in
  Natural Language Processing}, pages 1778--1783, Copenhagen, Denmark.
  Association for Computational Linguistics.

\bibitem[{Yang et~al.(2018)Yang, Qi, Zhang, Bengio, Cohen, Salakhutdinov, and
  Manning}]{DBLP:conf/emnlp/Yang0ZBCSM18}
Zhilin Yang, Peng Qi, Saizheng Zhang, Yoshua Bengio, William~W. Cohen, Ruslan
  Salakhutdinov, and Christopher~D. Manning. 2018.
\newblock \href {https://aclanthology.info/papers/D18-1259/d18-1259} {Hotpotqa:
  {A} dataset for diverse, explainable multi-hop question answering}.
\newblock In \emph{Proceedings of the 2018 Conference on Empirical Methods in
  Natural Language Processing, Brussels, Belgium, October 31 - November 4,
  2018}, pages 2369--2380.

\bibitem[{Yasunaga et~al.(2018)Yasunaga, Kasai, and
  Radev}]{yasunaga-etal-2018-robust}
Michihiro Yasunaga, Jungo Kasai, and Dragomir Radev. 2018.
\newblock \href {https://doi.org/10.18653/v1/N18-1089} {Robust multilingual
  part-of-speech tagging via adversarial training}.
\newblock In \emph{Proceedings of the 2018 Conference of the North {A}merican
  Chapter of the Association for Computational Linguistics: Human Language
  Technologies, Volume 1 (Long Papers)}, pages 976--986, New Orleans,
  Louisiana. Association for Computational Linguistics.

\bibitem[{Zhu et~al.(2020)Zhu, Cheng, Gan, Sun, Goldstein, and
  Liu}]{DBLP:freelb}
Chen Zhu, Yu~Cheng, Zhe Gan, Siqi Sun, Tom Goldstein, and Jingjing Liu. 2020.
\newblock \href {https://openreview.net/forum?id=BygzbyHFvB} {Freelb: Enhanced
  adversarial training for natural language understanding}.
\newblock In \emph{8th International Conference on Learning Representations,
  {ICLR} 2020, Addis Ababa, Ethiopia, April 26-30, 2020}. OpenReview.net.

\end{thebibliography}
\bibliographystyle{acl_natbib}

\end{document}